\documentclass[10pt,twocolumn,letterpaper]{article}

\usepackage[pagenumbers]{cvpr} 
\usepackage{algorithm}
\usepackage{amsfonts}
\usepackage{algpseudocode} 
\usepackage{booktabs}
\usepackage{multirow}
\usepackage[table]{xcolor}
\definecolor{cvprblue}{rgb}{0.21,0.49,0.74}
\usepackage[pagebackref,breaklinks,colorlinks,allcolors=cvprblue]{hyperref}

\def\paperID{LoViF-18} 
\def\confName{CVPR}
\def\confYear{2026}

\title{Don’t Waste Bits!\\ Adaptive KV-Cache Quantization for Lightweight On-Device LLMs}

\author{Sayed Pedram Haeri Boroujeni$^{*\dagger}$, Niloufar Mehrabi, Patrick Woods, Gabriel Hillesheim, Abolfazl Razi\\
Clemson University, USA\\
{\tt\small \{shaerib, nmehrab, pnwoods, ghilles, arazi\}@g.clemson.edu
}}

\begin{document}
\twocolumn[{
\maketitle

\vspace{-0.8cm}
\begin{center}
    \includegraphics[width=\textwidth,height=0.285\textheight]{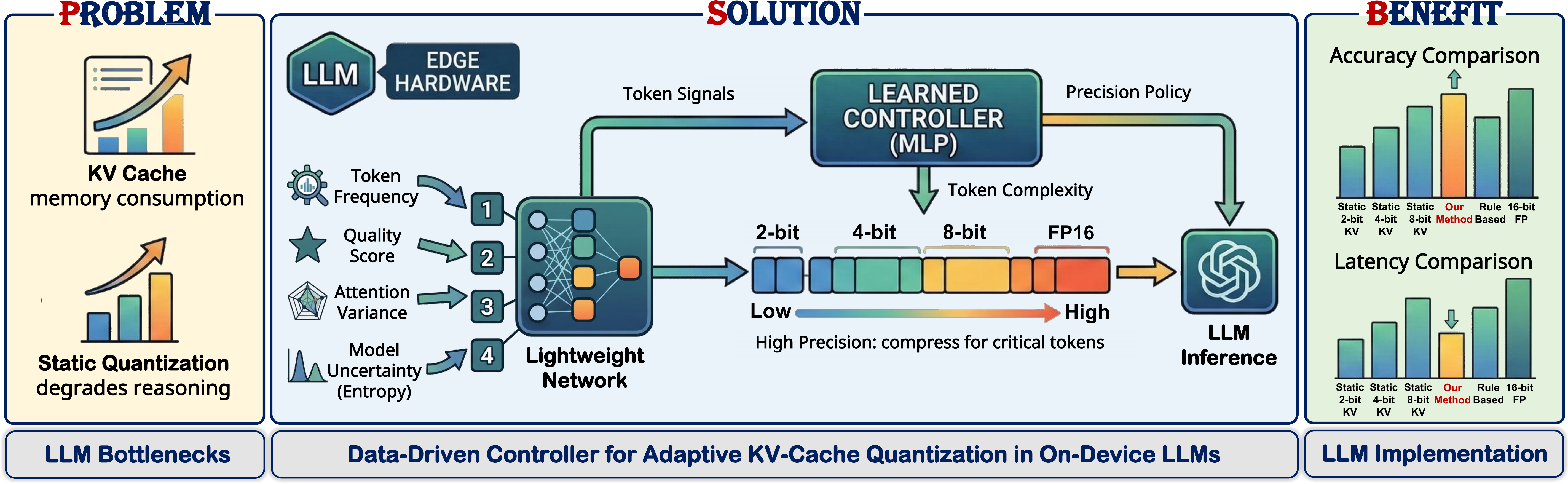}
    \vspace{-0.6cm}
    \captionof{figure}{\textbf{Overview of the proposed framework:} We introduce a data-driven controller for adaptive KV-cache quantization to address the KV-cache memory bottleneck in on-device LLM inference, where static quantization often degrades reasoning quality. Our method extracts lightweight token-level signals (e.g., token frequency, attention variance, and entropy-based uncertainty) and uses a learned MLP controller to assign per-token KV precision (2/4/8-bit or FP16) during decoding. This adaptive precision policy reduces KV memory footprint and latency while preserving (or improving) accuracy compared to static KV quantization, rule-based baselines, and FP16 inference.}

    \label{fig:overview}
\end{center}
}]

\begingroup
\renewcommand{\thefootnote}{\fnsymbol{footnote}}

\footnotetext[1]{Corresponding author: shaerib@g.clemson.edu} 
\footnotetext[2]{Project page: https://github.com/SayedPedramHaeri/Dont-Waste-Bits} 

\endgroup


\begin{abstract}
Large Language Models (LLMs) have achieved remarkable progress across reasoning, generation, and decision-making tasks, yet deploying them on mobile, embedded, and edge devices remains particularly challenging. On-device LLM inference is heavily constrained by the memory and bandwidth overhead of the key–value (KV) cache, which grows linearly with context length and often dominates decoding cost. Existing KV-cache quantization schemes typically rely on fixed precision or hand-crafted heuristics, thereby wasting bits on low-impact tokens while over-compressing informative ones, leading to avoidable accuracy degradation. Inspired by Huffman coding’s principle of variable-length allocation, we propose adaptive KV-cache quantization, a learned policy that assigns bit-width proportional to token importance, minimizing expected memory and latency without sacrificing competitive accuracy. Our framework extracts lightweight token-level features, including token frequency, quality score, attention variance, and entropy-based uncertainty, and feeds them into a compact data-driven controller that dynamically selects KV precision from \{2-bit, 4-bit, 8-bit, FP16\} during decoding. This adaptive precision policy reduces KV memory footprint and latency while improving accuracy compared to static KV quantization and rule-based baselines, and maintaining competitive accuracy close to FP16 inference across standard LLM benchmarks. Extensive experiments across multiple commonsense reasoning benchmarks using SmolLM-135M, SmolLM-360M, and SmolLM-1.7B demonstrate that our controller consistently improves the accuracy–latency trade-off. For instance, with SmolLM-360M on HellaSwag, our method reduces decoding latency (ms/token) by \textbf{17.75\%} relative to static KV quantization, improves accuracy by \textbf{7.60} points, and remains within only \textbf{0.30} points of FP16 inference.
\end{abstract}




\section{Introduction}
\label{sec:intro}


A primary bottleneck in autoregressive LLM decoding is the transformer KV cache \cite{vaswani2017attention, xing2025empowering, lu2025bluelm}, which stores the attention representations of previously processed tokens at each layer to avoid redundant computation during decoding \cite{hooper2024kvquant, he2024zipcache, cheng2025qaq, liu2024kivi}. Although this mechanism substantially improves computational reuse, it simultaneously introduces significant memory and bandwidth overhead that scales linearly with sequence length and accumulates across layers and heads \cite{zhang2023h2o, li2024snapkv}. As a result, long-context generation is often dominated by KV-cache storage, which becomes the dominant consumer of accelerator memory and a major source of latency, particularly on resource-constrained hardware with bandwidth-limited reads and writes \cite{wu2025scope, yu2025seqafford}. To address this challenge, improving KV-cache efficiency has emerged as an essential direction for on-device LLM deployment, motivating research on cache compression, quantization, and memory-aware decoding \cite{he2024zipcache, liu2024kivi, li2024snapkv, shu2025audio, boroujeni2026all}.

Existing KV-cache quantization methods typically rely on static precision assignments or hand-crafted heuristics to compress cached keys and values \cite{shutova2025cache, cheng2025qaq, hooper2024kvquant, li2024snapkv}. While these strategies are effective, they overlook the fact that tokens do not contribute equally to future predictions and that their sensitivity to quantization can vary significantly across the sequence \cite{haoyang2025survey, tao2025cocktail}. Uniformly allocating the same bit-width to all cached tokens can waste bandwidth on low-impact representations, while aggressively compressing informative tokens may disproportionately degrade model accuracy, leading to an unfavorable compression–performance trade-off \cite{zhou2024survey}. Moreover, heuristic rules often fail to generalize across model scales, datasets, and context lengths, making it difficult to maintain both efficiency and accuracy in practice across diverse deployment settings and evaluation conditions.\cite{zhou2024dynamickv, tao2025cocktail, lang2024comprehensive}.

This observation motivates a more principled view of KV-cache compression: precision should be treated as a limited resource and allocated selectively based on token importance \cite{adnan2024keyformer, tao2025moqae}. Inspired by variable-length allocation principles, an effective quantization strategy should assign more bits to informative tokens and fewer to less influential ones, thereby reducing memory usage and latency without unnecessarily sacrificing predictive quality \cite{buhlmann1999variable, liu2024minicache}. An adaptive allocation strategy is particularly well suited to lightweight and mid-scale LLMs deployed on-device, where even modest reductions in KV storage and data movement can yield meaningful gains in responsiveness and hardware efficiency \cite{feng2024ada, saxena2025resq}. More broadly, dynamic allocation provides a flexible alternative to static quantization and supports more favorable accuracy–efficiency trade-offs across diverse models and benchmarks \cite{shutova2025cache, he2024zipcache, likvtuner}.

This paper introduces \textit{\textbf{Don’t Waste Bits!}}, a data-driven framework for adaptive KV-cache quantization that predicts token importance and assigns token-wise precision during decoding accordingly. Inspired by Huffman coding’s principle of variable-length allocation, our method employs a compact controller that learns to dynamically assign KV precision from \{2-bit, 4-bit, 8-bit, FP16\} based on lightweight token-level features, reducing expected memory usage and latency while maintaining accuracy close to FP16 inference. Extensive experiments on commonsense reasoning benchmarks with SmolLM-135M, SmolLM-360M, and SmolLM-1.7B show that adaptive precision consistently achieves a better accuracy–latency trade-off than static KV quantization, rule-based baselines, and FP16 inference. The main contributions of this paper are summarized as follows:

\begin{itemize}
    \item \textbf{Adaptive KV precision allocation:} We introduce a token-wise KV-cache quantization policy that assigns bit-widths based on token importance during autoregressive decoding, thereby reducing KV overhead while bounding accuracy loss.

    \item \textbf{Lightweight data-driven controller:} We design a compact MLP controller for real-time quantization that leverages readily available token-level features, including token frequency or rarity, quality score, attention variance, and entropy-based uncertainty, to select KV precision under strict memory and latency constraints.

    \item \textbf{Comprehensive empirical evaluation:} We conduct a broad analysis of SmolLM-135M, SmolLM-360M, and SmolLM-1.7B across multiple commonsense reasoning benchmarks, showing improved accuracy over static and heuristic baselines while maintaining performance competitive with FP16 inference.

    \item \textbf{Efficient LLM deployment:} We further demonstrate the potential for deployment on mobile, embedded, and edge devices by consistently improving the trade-off among accuracy, latency, and KV-cache memory usage.
\end{itemize}

\section{Related Work}
\label{sec:Related}
Recent work on improving LLM inference efficiency has increasingly targeted the KV cache, whose storage and memory traffic become major bottlenecks in autoregressive decoding as context length grows. Existing approaches can be broadly categorized into (i) cache reduction strategies that reduce what must be stored or retrieved (e.g., selective retention, cache eviction, or recomputation); (ii) cache compression methods that shrink representation size through quantization, low-rank approximation, or structured compression; and (iii) systems-level optimizations that reorganize attention computation to better utilize hardware and memory hierarchies. These approaches differ in their assumptions about model internals, adaptation granularity, and trade-offs among accuracy, latency, and implementation complexity.

KV-cache quantization methods aim to reduce per-token KV memory while preserving attention fidelity. ZipCache \cite{he2024zipcache} combines an attention-based saliency metric with efficient mixed-precision quantization to reduce overhead while staying compatible with fast attention kernels. QAQ \cite{cheng2025qaq} shows that keys and values have different quantization sensitivities and introduces non-uniform, attention-aware strategies with outlier handling to achieve high compression with minimal quality loss. KVQuant \cite{hooper2024kvquant} further enables ultra-long-context inference through tailored designs, including per-channel and pre-RoPE Key quantization, dense-and-sparse per-vector quantization, and non-uniform KV quantization, achieving sub-4-bit compression with only minor perplexity degradation.

KV-cache pruning and retention methods reduce memory and computation by keeping only the most important cached tokens, motivated by the sparsity of attention in long contexts. H2O \cite{zhang2023h2o} shows that a small set of heavy-hitter tokens dominates attention and proposes an eviction policy that preserves both recent and high-impact tokens. Keyformer \cite{adnan2024keyformer} similarly identifies key tokens during inference and retains only them in the cache, substantially reducing KV size and bandwidth while maintaining accuracy. SnapKV \cite{li2024snapkv} extends this idea to the attention-head level, using prompt observations to predict salient KV positions and compressing the cache by selecting clustered important tokens for each head.

Adaptive cache management and complementary compression improve efficiency by controlling how KV states are retained while also reducing model weight cost. FastGen \cite{ge2023model} profiles attention patterns across heads and adaptively retains or evicts KV content based on head structure, reducing memory with negligible quality loss. Ada-KV \cite{feng2024ada} shows that uniform compression across heads is suboptimal and instead allocates eviction budgets adaptively using a theoretical loss bound. Complementary to KV-focused methods, AWQ \cite{lin2024awq} reduces the model weight footprint through weight-only quantization, protecting a small set of salient weights to lower quantization error and enable hardware-friendly low-bit deployment. Because it targets weights rather than the KV cache, AWQ is largely orthogonal to KV-cache optimization and can be combined with KV-focused methods.

Previous studies reduce KV-cache cost through token retention, numerical compression, and weight-only quantization. However, many still rely on static precision assignments or hand-crafted rules that ignore token-level importance, while retention-based methods may remove information critical to downstream accuracy. This motivates adaptive KV policies that allocate precision selectively, preserving informative tokens while more aggressively compressing low-impact ones. Accordingly, we introduce a lightweight data-driven controller for token-wise KV precision allocation during decoding, improving the accuracy-latency trade-off for on-device deployment.

\section{Methodology}
\label{sec:Methodology}
In this section, we present the proposed adaptive KV-cache quantization framework for efficient on-device LLM inference. We begin by formalizing the problem (Section \ref{sec:problem}) and reviewing the attention and decoding preliminaries relevant to KV-cache construction and quantization (Section \ref{sec:preliminary}). We then introduce the theoretical intuition behind our approach, viewing KV precision as a limited resource that should be allocated selectively according to token importance (Section \ref{sec:theoretical}). Motivated by this perspective, we propose \textit{\textbf{Don’t Waste Bits!}}, an adaptive KV-cache quantization framework for token-wise precision allocation (Section \ref{sec:overview}). We describe the end-to-end pipeline, including the token-level saliency features used for importance estimation (Section \ref{sec:saliency}), the lightweight controller network that predicts precision assignments (Section \ref{sec:controller}), the adaptive KV-cache quantization mechanism (Section \ref{sec:quantization}), and the training objective (Section \ref{sec:training}). 

\subsection{Problem Description}
\label{sec:problem}
In autoregressive language modeling, the goal is to estimate the probability of the next token given a prefix $x_{1:n} = \{x_1, x_2, \ldots, x_n\}$, i.e., $p(x_{n+1} \mid x_{1:n})$, and to generate text by repeatedly sampling or selecting the next token according to this distribution. In generative LLMs, inference typically proceeds in two stages:

\textbf{(1) Prompt Encoding Phase:} The input context is processed once under causal masking to compute attention activations for all prompt tokens. During this stage, each transformer layer produces the corresponding key and value tensors for every token, which are then stored for later reuse. 

\textbf{(2) Token Generation Phase:} During this stage, the model generates new tokens sequentially in an autoregressive manner. At each step, the newly generated token is passed through all layers, and self-attention attends over the full history of previously processed tokens.

A key mechanism behind efficient autoregressive decoding is the KV cache, which stores per-layer keys and values for previously processed tokens. Without it, the model would need to recompute $K$ and $V$ for the entire prefix at every decoding step, resulting in substantial redundant computation. Instead, the cache is built incrementally by storing prompt tokens during encoding and appending each newly generated token during decoding. Although this significantly reduces computation, it introduces a major inference bottleneck on resource-constrained hardware, since KV storage and memory traffic grow linearly with context length and accumulate across layers.

\subsection{Preliminary}
\label{sec:preliminary}
Building on the above discussion, we briefly review the attention formulation and KV-cache construction in autoregressive LLM inference, since they form the main computational bottleneck targeted by our method. Consider an input sequence of token embeddings $X \in \mathbb{R}^{l \times d_{\mathrm{model}}}$, where $l$ denotes the sequence length and $d_{\mathrm{model}}$ is the hidden dimension. In a standard self-attention block with projection matrices $W_Q$, $W_K$, and $W_V$, the query, key, and value tensors are computed during prompt encoding as:

\begin{equation}
Q = XW_Q, \quad K = XW_K, \quad V = XW_V
\end{equation}

\noindent The corresponding attention output is computed from the scaled dot-product attention mechanism:

\begin{equation}
\mathrm{Attn}(Q,K,V) = \mathrm{Softmax}\left(\frac{QK^\top}{\sqrt{d_k}}\right)V
\end{equation}

\noindent where $d_k$ denotes the key dimension. In autoregressive inference, the key and value tensors are stored in memory as the KV cache for subsequent decoding steps. 
During the token generation phase, decoding proceeds one step at a time. Let $x \in \mathbb{R}^{d_{\mathrm{model}}}$ denote the embedding of the current token. The query for the current step is computed as follows:

\begin{equation}
q = xW_Q
\end{equation}

\noindent Meanwhile, the corresponding $K$ and $V$ vectors for the current token are then appended to the existing cache tensors:

\begin{equation}
K \leftarrow \mathrm{Concat}(K, xW_K), \quad V \leftarrow \mathrm{Concat}(V, xW_V)
\end{equation}

\noindent The attention output for the current step is then given by:

\begin{equation}
a = \mathrm{Softmax}\left(\frac{qK^\top}{\sqrt{d_k}}\right)V
\end{equation}

This formulation highlights the central challenge addressed in our paper. Although each decoding step processes only one new token, it must repeatedly access an ever-growing KV cache, causing both memory usage and memory traffic to scale with context length. For clarity and consistency throughout the remainder of the paper, we use $b$ to denote the batch size, $h$ the number of attention heads, $l$ the sequence length, and $d$ the per-head feature dimension.

\subsection{Theoretical Foundation}
\label{sec:theoretical}
Our adaptive KV-cache quantization framework is inspired by Huffman’s Optimality Theorem \cite{huffman1952}, assigning bit-widths in proportion to token importance. In particular, the expected code length $\mathbb{E}[\ell(x)]$ is minimized when more probable symbols receive shorter codes and less probable symbols receive longer ones:

\begin{equation}
    \ell^{*}(x) = \lceil -\log_{2} p(x) \rceil
    \label{eq:huffman_optimal}
\end{equation}

\noindent where $p(x)$ denotes the probability (frequency) of symbol $x$. Therefore, common and predictable symbols receive shorter codewords, whereas rare and more informative symbols receive longer codewords.

In the context of LLMs, we posit that tokens $x \in \mathcal{X}$ do not contribute equally to the final hidden state or the attention output. To capture this variability, we define a token-importance function $I(x) \in [0,1]$ based on contextual features such as attention entropy, activation magnitude, and positional index. By analogy to~\eqref{eq:huffman_optimal}, the optimal bit-width assignment $b^*(x)$ should satisfy:

\begin{equation}
    b^{*}(x) = f(I(x))
    \label{eq:adaptive_bitwidth}
\end{equation}

\noindent where $f: [0,1] \to \mathcal{B}$ maps token importance to a quantization class $b \in \mathcal{B} = \{2, 4, 8, 16\}$.  The central principle is that tokens with lower importance contain less task-relevant information and can therefore be represented with fewer bits. Since $I(x)$ is not directly observable a priori, we approximate the composite mapping $f \circ I$ with a neural controller $f_{\theta}: \mathcal{X} \to \mathcal{B}$, trained end-to-end to minimize expected latency while constraining accuracy degradation.

Following this perspective, we frame the controller's task as a constrained optimization problem. Let $Q(x, b)$ denote the quality (accuracy) associated with assigning token $x$ to bit-width $b$, and let $L(b)$ denote the corresponding latency cost. The goal of the controller is to maximize the expected fitness $F(\theta)$ as follows:

\begin{equation}
    \max_{\theta} \mathbb{E}_{x \sim \mathcal{X}} \left[ U(Q(x, f_\theta(x))) - \lambda \cdot K(L(f_\theta(x))) \right]
\end{equation}

\noindent where $U(\cdot)$ is a utility function, $K(\cdot)$ is a cost function, and $\lambda$ is a Lagrange multiplier that controls the trade-off between accuracy and latency. Let $\mathcal{S}_{\mathrm{fixed}}$ denote a system with a uniform bit-width $b_{\mathrm{fixed}} = 16$. We say that $\mathcal{S}_{\mathrm{adapt}}$ Pareto-dominates $\mathcal{S}_{\mathrm{fixed}}$ if it achieves lower expected latency while incurring at most a bounded accuracy deviation:

\begin{equation}
\begin{aligned}
\mathbb{E}[L(\mathcal{S}_{\mathrm{adapt}})] &< \mathbb{E}[L(\mathcal{S}_{\mathrm{fixed}})], \\
|A_{\mathrm{fixed}} - A_{\mathrm{adapt}}| &< \epsilon .
\end{aligned}
\end{equation}

\noindent For any token assigned low importance under $I(x)$, the controller selects a bit-width $b < 16$. Since $L(b) < L(16)$ for all $b \in \{2,4,8\}$, the adaptive policy reduces total latency:

\begin{equation}
    \Delta L = \sum_{x \in \mathcal{X}_{low}} p(x) [L(16) - L(f_\theta(x))] > 0
\end{equation}

\noindent where $\mathcal{X}_{\mathrm{low}}$ denotes the set of tokens identified by the controller as having low contextual importance. This result highlights that the adaptive scheme offers a structural improvement over fixed-precision baselines by reducing the informational waste inherent in uniform quantization.

To further support this claim, we evaluate performance through the expected bit-width $\mathbb{E}[b]$ and total distortion $D$. In a fixed-precision system, resource consumption is constant: $\mathbb{E}[b]_{\mathrm{fixed}} = 16$. Conversely, in our adaptive system:

\begin{equation}
    \mathbb{E}[b]_\mathrm{adaptive} = \sum_{x \in \mathcal{X}} p(x) \cdot f_\theta(x)
\end{equation}

\noindent According to the Source Coding Theorem~\cite{shannon1948mathematical}, the most efficient representation is achieved when resource allocation is matched to the underlying information content (entropy). We define $H(x)$ as the \textit{quantization entropy}, the minimum number of bits required to represent token $x$ while preserving its contribution to the attention output. Our system minimizes the computational ``waste'' $W = \mathbb{E}[b] - H(X)$, which cannot be eliminated by a fixed-rate encoder.

The improvement $\Delta$ can be interpreted as moving closer to the \textit{rate--distortion bound} $R(D)$, defined as:

\begin{equation}
    R(D) = \min_{p(\hat{x}|x): \sum p(x, \hat{x})d(x, \hat{x}) \leq D} I(X; \hat{X})
\end{equation}

\noindent where $I(X;\hat{X})$ denotes the \textit{mutual information} between the original token $X$ and its quantized representation $\hat{X}$, corresponding to the minimum information rate required to achieve distortion $D$. By allowing $b$ to vary with $I(x)$, the controller allocates higher precision to tokens with greater contextual importance. Consequently, under a fixed latency budget $\mathcal{L}$, the adaptive policy is better positioned to preserve accuracy than a fixed-precision baseline. This selective allocation improves the information-to-latency trade-off:

\begin{equation}
    \left.\frac{\partial A}{\partial L}\right|_{\mathrm{adaptive}} >
    \left.\frac{\partial A}{\partial L}\right|_{\mathrm{fixed}} 
\end{equation}

Taken together, this theoretical foundation motivates our adaptive controller as an effective approach to the Pareto-optimal accuracy--efficiency frontier in LLM inference.

\subsection{Framework Overview}
\label{sec:overview}
Figure \ref{fig:overview} provides an overview of the proposed framework, \textit{\textbf{Don’t Waste Bits!}}, a data-driven adaptive KV-cache quantization method for efficient on-device LLM inference. Rather than assigning a uniform KV precision to all cached tokens, the framework learns a token-level policy that dynamically selects the precision for each token’s key and value representations during decoding. Based on lightweight contextual features, it estimates each token’s relative importance and allocates precision accordingly. As a result, the method reduces unnecessary memory use for low-impact tokens while preserving higher precision for tokens more likely to influence future predictions.

Conceptually, the proposed pipeline consists of four stages. First, for each token during decoding, we extract a small set of lightweight saliency features that capture token importance, uncertainty, and contextual influence, while remaining computationally efficient on resource-constrained hardware. Second, these features are passed to a lightweight controller network, implemented as a compact multi-layer perceptron (MLP), which predicts a precision class for each token. The controller selects one of four candidate KV storage levels, namely $\{2,4,8,16\}$ bits, where 16-bit corresponds to FP16 storage. Third, the selected precision is used to quantize the token’s key and value tensors before they are appended to the KV cache. Finally, the quantized cache is used in subsequent decoding steps as in standard autoregressive inference, except that tokens are stored at heterogeneous precision levels rather than a fixed bit-width.


\subsection{Token-Level Saliency Features}
\label{sec:saliency}

One of the key components of the proposed framework is the extraction of lightweight token-level saliency features used to estimate each token’s contextual importance during decoding. Instead of relying on expensive auxiliary models or deep architectural modifications, our  method derives a small set of inexpensive features directly from the model’s forward pass. These features capture complementary aspects of token behavior, including predictive uncertainty, contextual rarity, and attention dynamics, making them informative for selecting the appropriate KV precision. Intuitively, tokens that are more uncertain, less predictable, or more structurally influential are more likely to require higher-precision KV representations, whereas stable and low-impact tokens can often be stored at lower bit-widths.

Specifically, for each token $x_t$, we compute three saliency features: \emph{entropy}, \emph{rarity}, and \emph{attention variance}. Let $z_t \in \mathbb{R}^{|\mathcal{V}|}$ denote the pre-softmax logits at position $t$, where $\mathcal{V}$ is the vocabulary. We first quantify predictive uncertainty through the entropy of the next-token distribution:

\begin{equation}
H_t = - \sum_{v \in \mathcal{V}} p_t(v)\log p_t(v),
\end{equation}

\noindent where $p_t(v) = \mathrm{Softmax}(z_t)_v$. Higher entropy indicates greater uncertainty in the model’s predictive distribution, suggesting that the corresponding token may be more sensitive to compression. Accordingly, we use entropy as a proxy for the degree of representational fidelity that should be preserved in the KV cache.

The second feature measures token rarity. Let $c(x_t)$ denote the running count of token $x_t$ in the training stream, and let $N$ denote the total number of observed tokens.  We define rarity through a smoothed self-information score:

\begin{equation}
R_t = -\log \left( \frac{c(x_t)+1}{N + |\mathcal{V}_{\mathrm{obs}}| + 1} \right)
\end{equation}

\noindent where $|\mathcal{V}_{\mathrm{obs}}|$ is the number of distinct observed tokens. This measure assigns larger values to infrequent or less predictable tokens and smaller values to common tokens. The intuition is that rare tokens often carry more specific semantic information and may therefore require more careful preservation under quantization.

The third signal captures variability in the attention pattern. Let $A^{(L)} \in \mathbb{R}^{h \times l \times l}$ denote the attention tensor from the final transformer layer, where $h$ is the number of attention heads and $l$ is the sequence length. We compute the attention-variance feature as follows:

\begin{equation}
V_t = \frac{1}{h}\sum_{i=1}^{h} \mathrm{Var}\!\left(A^{(L)}_i\right)
\end{equation}

\noindent where $V_t$ reflects the sharpness or unevenness of the attention distribution and serves as a coarse indicator of structural sensitivity in the current context. Tokens associated with more variable attention patterns may engage in less uniform, more context-dependent interactions, suggesting they may benefit from higher-precision preservation.

These three features, together with an additional token-level confidence term ($C_t$), provide complementary information. Entropy captures predictive uncertainty, rarity reflects token-level informativeness, attention variance provides a lightweight measure of contextual structure, and confidence reflects the model’s certainty in its prediction. Collectively, they form a compact feature vector:

\begin{equation}
s_t = [H_t,\; R_t,\; V_t, \; C_t] \in \mathbb{R}^4
\end{equation}

\noindent which is then passed to the controller network for precision prediction. In our implementation, these features are extracted from benchmark contexts used to build the controller training set, and each token is paired with its feature vector and associated precision and latency labels.

\subsection{Lightweight Controller Network}
\label{sec:controller}
To translate token-level importance features into KV-precision decisions, we introduce a lightweight controller network that predicts the bit-width assigned to each token during decoding. The controller is compact, fast, and easy to integrate into autoregressive inference, ensuring negligible computational overhead while achieving significant memory and latency savings through adaptive quantization. To satisfy the constraints of on-device deployment, we intentionally avoid heavy auxiliary modules and instead adopt a shallow MLP that operates on a low-dimensional feature vector for each token.

For each token, the controller receives the feature vector ($s_t$) introduced in Section \ref{sec:saliency}. The controller maps $s_t$ to one of four discrete precision classes corresponding to:
\begin{equation}
B = \{2,4,8,16\}
\end{equation}

\noindent where 16 denotes FP16 storage. Concretely, we use a three-layer MLP with two hidden layers and ReLU nonlinearities:

\begin{equation}
h_t^{(1)} = \mathrm{ReLU}(W_1 \mathbf{s}_t + b_1)
\end{equation}
\begin{equation}
h_t^{(2)} = \mathrm{ReLU}(W_2 \mathbf{h}_t^{(1)} + b_2)
\end{equation}
\begin{equation}
o_t = W_3 \mathbf{h}_t^{(2)} + b_3
\end{equation}

\noindent Here, $W_1$, $W_2$, and $W_3$ are the learnable weight matrices of the three linear layers, and $b_1$, $b_2$, and $b_3$ are the corresponding bias vectors. The vectors $\mathbf{h}_t^{(1)}$ and $\mathbf{h}_t^{(2)}$ denote the hidden representations produced by the first and second hidden layers, respectively, after applying the ReLU activation. The output vector $\mathbf{o}_t \in \mathbb{R}^{4}$ contains the logits over the four candidate precision classes.

A key property of the controller is that it predicts a \emph{distribution} over bit-width classes rather than making hard decisions during optimization. Specifically, the predicted class probabilities are given by:

\begin{equation}
p_t = \mathrm{Softmax}(\mathbf{o}_t)
\end{equation}

\noindent where $p_t \in \mathbb{R}^{4}$ denotes the probability distribution over the four candidate precision classes. The final predicted bit-width is then obtained as follows:

\begin{equation}
\hat{b}_t = \mathrm{IndexToBit}\!\left(\arg\max_{k \in \{1,2,3,4\}} p_{t,k}\right)
\end{equation}

These probabilities allow the training objective to model differentiable expectations of latency and quality, enabling the controller to balance classification accuracy with efficiency and predictive fidelity. The controller therefore serves not only as a quantization-level classifier, but also as a lightweight decision module that allocates KV precision according to token importance, latency cost, and quality preservation. We use a hidden dimension of 128 to maintain expressiveness with low inference overhead. Training is performed on token-level samples from benchmark contexts, each consisting of a feature vector, a target precision label, and measured latency and quality statistics. Targets are mapped to four classes corresponding to 2-bit, 4-bit, 8-bit, and 16-bit KV storage, and an 80/20 stratified train-validation split is used to preserve class balance.



\subsection{Adaptive KV Quantization}
\label{sec:quantization}
Given the controller prediction $\hat{b}_t \in \mathcal{B}$ for token $x_t$, we quantize the corresponding key and value tensors before appending them to the KV cache. Let $k_t$ and $v_t$ denote the key and value representations generated for token $x_t$ at a given layer. The controller-selected bit-width determines the precision used to store these tensors:
\begin{equation}
\hat{k}_t = Q_{\hat{b}_t}(\mathbf{k}_t), \qquad \hat{v}_t = Q_{\hat{b}_t}(\mathbf{v}_t).
\end{equation}
where $Q_{\hat{b}_t}(\cdot)$ denotes quantization under the assigned bit-width $\hat{b}_t \in \{2,4,8,16\}$. The quantized tensors $(\hat{k}_t,\hat{v}_t)$ are then appended to the cache and reused in decoding steps.

Unlike fixed-precision baselines, which assign a uniform bit-width to all cached tokens, our method enables heterogeneous precision allocation across the sequence. Tokens estimated to be less important are stored at lower precision to reduce KV-cache memory usage and memory traffic, whereas more important tokens retain higher precision to mitigate harmful information loss. In this way, adaptive KV quantization translates token-level controller decisions into a practical inference mechanism that improves the trade-off among memory usage, latency, and predictive performance.

\subsection{Training Objective}
\label{sec:training}

The controller is trained to balance accurate precision prediction with efficient inference behavior. Given controller logits $o_t$ and the corresponding class probabilities $p_t=\mathrm{Softmax}(o_t)$, we first use a standard cross-entropy loss to supervise the predicted precision class:

\begin{equation}
\mathcal{L}_{\mathrm{ce}} = \mathrm{CrossEntropy}(o_t, y_t)
\end{equation}
where $y_t$ is the target bit-width label. To explicitly encourage efficient decisions, we additionally incorporate an expected latency term as follows: 
\begin{equation}
\mathcal{L}_{\mathrm{lat}} = \sum_{k=1}^{4} p_{t,k} c_k
\end{equation}
where $c_k$ denotes the latency cost associated with class $k$, and an expected quality penalty as follows:
\begin{equation}
\mathcal{L}_{\mathrm{qual}} = 1 - \sum_{k=1}^{4} p_{t,k} q_k
\end{equation}
where $q_k$ denotes the class-wise quality score estimated from the training data. The final objective combines these three terms as follows:
\begin{equation}
\mathcal{L} = \alpha \mathcal{L}_{\mathrm{ce}} + \beta \mathcal{L}_{\mathrm{lat}} + \gamma \mathcal{L}_{\mathrm{qual}}
\end{equation}
where $\alpha$, $\beta$, and $\gamma$ control the trade-off among classification accuracy, latency reduction, and quality preservation. This objective encourages the controller to produce precision assignments that are not only label-consistent but also effective for improving the overall accuracy--efficiency trade-off during inference. Eventually, Algorithm \ref{alg:controller} summarizes the training and inference procedures of \textit{\textbf{Don’t Waste Bits!}}.

\begin{algorithm}[h]
\footnotesize
\caption{\small Don’t Waste Bits!}
\label{alg:controller}
\begin{algorithmic}[1]
\State \textbf{Input:} Token-level dataset $\mathcal{D}=\{(s_t, y_t)\}_{t=1}^{N}$
\State \textbf{Input:} Latency cost vector $\mathbf{c}\in\mathbb{R}^{4}$, loss weights $\alpha,\beta,\gamma$
\State \textbf{Output:} Trained controller $f_{\theta}$

\State Map KV bit-width labels $\{2,4,8,16\}$ to class indices $\{0,1,2,3\}$
\State Split $\mathcal{D}$ into stratified training and validation sets
\State Initialize controller $f_{\theta}$ as an MLP with two ReLU hidden layers
\State Estimate class-wise quality scores $\mathbf{q}\in\mathbb{R}^{4}$ from the training set

\For{each epoch}
    \For{each mini-batch $(\mathbf{Z}, \mathbf{y})$}
        \State Compute logits: $O \gets f_{\theta}(\mathbf{Z})$
        \State Compute class probabilities: $P \gets \mathrm{Softmax}(\mathbf{O})$
        \State Compute classification loss: $\mathcal{L}_{\mathrm{ce}} \gets \mathrm{CrossEntropy}(\mathbf{O}, \mathbf{y})$
        \State Compute expected latency: $\mathcal{L}_{\mathrm{lat}} \gets \frac{1}{B}\sum_{i=1}^{B}\sum_{k=1}^{4} P_{ik}c_k$
        \State Compute expected quality: $Q_{\mathrm{exp}} \gets \frac{1}{B}\sum_{i=1}^{B}\sum_{k=1}^{4} P_{ik}q_k$
        \State Compute quality penalty: $\mathcal{L}_{\mathrm{qual}} \gets 1 - Q_{\mathrm{exp}}$
        \State Compute total loss: $\mathcal{L} \gets \alpha \mathcal{L}_{\mathrm{ce}} + \beta \mathcal{L}_{\mathrm{lat}} + \gamma \mathcal{L}_{\mathrm{qual}}$
        \State Update $\theta$ using Adam
    \EndFor
\EndFor

\State \Return $f_{\theta}$
\end{algorithmic}
\end{algorithm}
\section{Experiments}
In this section, we evaluate the proposed adaptive KV-cache quantization framework on multiple language models and commonsense reasoning benchmarks. We first describe the experimental settings, then present quantitative results to assess the trade-off between predictive accuracy and decoding latency. All experiments are conducted on an NVIDIA RTX 4090 GPU with 24\,GB of memory.

\subsection{Experimental Settings}

\noindent \textbf{Backbones:} We employ three open-source SmolLM base models, namely SmolLM-135M, SmolLM-360M, and SmolLM-1.7B, to evaluate the proposed framework across small, moderate, and relatively larger parameter scales. The detailed architecture specifications of the adopted backbones are summarized in Table \ref{tab:backbone_models}.

\begin{table}[h]
\centering
\caption{Architecture of SmolLM models used in our experiments.}
\vspace{-0.2cm}
\label{tab:backbone_models}
\setlength{\tabcolsep}{2.8pt}
\renewcommand{\arraystretch}{1.0}
\resizebox{\columnwidth}{!}{%
\begin{tabular}{lcccccc}
\toprule
Model & \#Param & \#Layers & \#Head & Hidden Dim & LR & BS \\
\midrule
SmolLM-135M & 135M  & 30 & 9  & 1536 & 3e-3 & 1M \\
SmolLM-360M & 362M  & 32 & 15 & 2560 & 3e-3 & 1M \\
SmolLM-1.7B & 1.71B & 24 & 32 & 8192 & 5e-4 & 2M \\
\bottomrule
\end{tabular}%
}
\end{table}

\noindent \textbf{Datasets:} We evaluate the proposed framework on three challenging benchmarks: HellaSwag \cite{zellers2019hellaswag}, OpenBookQA (OBQA) \cite{OpenBookQA2018}, and ARC-Challenge \cite{clark2018think}. Together, these benchmarks provide a diverse testbed for evaluating the proposed method across different reasoning settings. A summary of the evaluation datasets is provided in Table \ref{tab:datasets}.

\begin{table}[h]
\centering
\caption{Statistics of language datasets used in our experiments.}
\vspace{-0.2cm}
\label{tab:datasets}
\setlength{\tabcolsep}{2.8pt}
\renewcommand{\arraystretch}{1.0}
\resizebox{\columnwidth}{!}{%
\begin{tabular}{llccc}
\toprule
Dataset &Category  & Train & Validation & Test \\
\midrule
HellaSwag &Commonsense Reasoning   & 39,905 & 10,042 & 10,003 \\
ARC-C      &Science Reasoning  & 99,842 & 1,531  & 14,042 \\
OBQA      &Science QA  & 4,957  & 500    & 500 \\
\bottomrule
\end{tabular}%
}
\end{table}


\noindent \textbf{Metrics:} We report two primary evaluation metrics: \emph{accuracy} and \emph{latency}. Accuracy is computed based on the final multiple-choice answer selected by the model, while latency is measured in milliseconds per token (ms/token) and captures the average time required to generate each token under different KV-cache quantization policies.

\noindent \textbf{Baselines:} We compare the proposed method against three KV-cache precision baselines: FP16 inference without quantization, static 4-bit KV quantization, and a rule-based dynamic KV policy. FP16 serves as the full-precision reference, while the static baselines apply a uniform bit-width to all cached tokens, and the rule-based method uses hand-crafted heuristics. Additionally, we evaluate the proposed method across modern LLM families, including Pythia \cite{biderman2023pythia}, Cerebras-GPT \cite{dey2023cerebras}, LaMini-GPT \cite{wu2024lamini}, Galactica \cite{taylor2022galactica}, and OPT \cite{zhang2022opt}, to ensure a broad evaluation setting.

\subsection{Experimental Results}
To evaluate the effectiveness and deployability of the proposed framework, we compare it against multiple KV-cache precision baselines, with the results summarized in Table \ref{tab:baselines}. Furthermore, Table \ref{tab:models} provides a comparison of the proposed method with representative LLMs. All methods are evaluated in the zero-shot setting on downstream reasoning benchmarks, where the reported accuracy of existing methods are directly taken from \cite{dey2023cerebras}. For fair comparison, we evaluate our model under the same experimental settings.




\newcommand{\myrowcolour}{\rowcolor[gray]{0.9}}
\begin{table}[h]
\centering
\caption{Comparison of the proposed method with baselines.}
\vspace{-0.2cm}
\label{tab:baselines}
\setlength{\tabcolsep}{2.8pt}
\renewcommand{\arraystretch}{1.0}
\resizebox{\columnwidth}{!}{%
\begin{tabular}{llccc}
\toprule
\multicolumn{5}{c}{\textbf{Number of Parameters $=$ SmolLM-135M}} \\ 
\midrule
Baselines & Metrics & HellaSwag & OBQA & ARC-C \\
\midrule
\multirow[t]{2}{*}{FP16}                    & Accuracy (\%) &37.20  &29.00  &28.20  \\
                                            & Latency  ($s/token$) &3.45  &1.97  &2.57  \\
\multirow[t]{2}{*}{Static 4-bit KV}         & Accuracy (\%) &33.60  &27.60  &25.00  \\
                                            & Latency  ($s/token$) &2.93  &1.72 &2.25  \\
\multirow[t]{2}{*}{Dynamic KV}              & Accuracy (\%) &29.90  &23.80  &25.50 \\
                                            & Latency  ($s/token$) &3.21  &1.59  &2.20  \\
\myrowcolour
\multirow[t]{2}{*}{\textbf{Ours}}                    & Accuracy (\%) &\textbf{36.90}  &\textbf{28.60}  &\textbf{27.00}  \\
\myrowcolour
                                            & Latency  ($s/token$) &\textbf{2.25}   &\textbf{1.00}  &\textbf{2.08}  \\
\bottomrule

\multicolumn{5}{c}{\textbf{Number of Parameters $=$ SmolLM-360M}} \\ 
\midrule
Baselines & Metrics & HellaSwag & OBQA & ARC-C \\
\midrule
\multirow[t]{2}{*}{FP16}                    & Accuracy (\%) &41.50  &31.00  &32.00  \\
                                            & Latency  ($s/token$) &3.50  &2.20  &2.90  \\
\multirow[t]{2}{*}{Static 4-bit KV}         & Accuracy (\%) &33.60  &30.40  &30.00  \\
                                            & Latency  ($s/token$) &2.93  &1.81  &2.83  \\
\multirow[t]{2}{*}{Dynamic KV}              & Accuracy (\%) &29.90  &22.00  &23.50  \\
                                            & Latency  ($s/token$) &3.21  &1.60  &2.70  \\
\myrowcolour
\multirow[t]{2}{*}{\textbf{Ours}}                    & Accuracy (\%) &\textbf{41.20}  &\textbf{30.50}  &\textbf{31.50}  \\
\myrowcolour
                                            & Latency  ($s/token$) &\textbf{2.41}   &\textbf{1.40}  &\textbf{2.43}  \\
\bottomrule

\multicolumn{5}{c}{\textbf{Number of Parameters $=$ SmolLM-1.7B}} \\ 
\midrule
Baselines & Metrics & HellaSwag & OBQA & ARC-C \\
\midrule
\multirow[t]{2}{*}{FP16}                    & Accuracy (\%) &49.00  &33.00  &33.50  \\
                                            & Latency  ($s/token$) &2.15  &1.50  &2.45  \\
\multirow[t]{2}{*}{Static 4-bit KV}         & Accuracy (\%) &41.10  &30.00  &32.30  \\
                                            & Latency  ($s/token$) &1.89  &1.10  &2.15  \\
\multirow[t]{2}{*}{Dynamic KV}              & Accuracy (\%) &27.50  &25.00  &26.00  \\
                                            & Latency  ($s/token$) &1.96  &1.25  &2.21  \\
\myrowcolour
\multirow[t]{2}{*}{\textbf{Ours}}                    & Accuracy (\%) &\textbf{48.60}  &\textbf{31.60}  &\textbf{33.00}  \\
\myrowcolour
                                            & Latency  ($s/token$) &\textbf{1.58}   &\textbf{1.00}  &\textbf{1.90}  \\
\bottomrule
\end{tabular}%
}
\end{table}

\begin{table}[h]
\centering
\caption{Comparison of the proposed method with LLMs.}
\vspace{-0.2cm}
\label{tab:models}
\setlength{\tabcolsep}{2.8pt}
\renewcommand{\arraystretch}{1.0}
\resizebox{\columnwidth}{!}{%
\begin{tabular}{lccccc}
\toprule
\multicolumn{5}{c}{\textbf{Number of Parameters $\approx$ 135M}} \\ 
\midrule
Models & HellaSwag & OBQA & ARC-C & Average\\
\midrule
OPT-125M &29.20  &16.60  &18.90    &21.57\\
Pythia-70M &27.00  &13.20  &18.50    &19.57\\
Pythia-160M &29.30  &16.00  &18.10    &21.13\\
Galactica-125M &29.60  &28.20  &26.20    &28.00  \\
LaMini-GPT-124M &30.20  &29.60  &26.00    &28.60  \\
Cerebras-GPT-111M &26.80  &11.80  &16.60    &18.40\\
Cerebras-GPT-256M &27.40  &15.80  &17.00    &20.00\\
\myrowcolour
\textbf{SmolLM-135M $+$ Ours} &\textbf{36.90}  &\textbf{28.60} &\textbf{27.00}    &\textbf{30.83}\\
\bottomrule
\multicolumn{5}{c}{\textbf{Number of Parameters $\approx$ 360M}} \\ 
\midrule
Models & HellaSwag & OBQA & ARC-C & Average\\
\midrule
OPT-350M  &32.00  &17.60  &20.70    &23.43\\
Pythia-410M &33.30  &17.80  &21.30   &24.13\\
Pythia-1B &37.60  &19.60  &24.30    &27.17\\
Cerebras-GPT-590M   &29.10  &15.80  &19.00    &21.30\\
\myrowcolour
\textbf{SmolLM-360M $+$ Ours} &\textbf{41.20}  &\textbf{30.50}  &\textbf{31.50}    & \textbf{34.40}\\
\bottomrule
\multicolumn{5}{c}{\textbf{Number of Parameters $\approx$ 1.7BM}}\\ 
\midrule
Models & HellaSwag & OBQA & ARC-C & Average\\
\midrule
OPT-1.3B &41.50  &23.40  &23.40    &29.43\\
OPT-2.7B &45.80  &25.00  &26.80    &32.53\\
Pythia-1.4B  &39.80  &22.40  &25.60    &29.27\\
Pythia-2.8B  &45.10  &22.00  &28.80   &31.97\\
Cerebras-GPT-1.3B  &32.50  &16.60  &22.40    &23.83\\
Cerebras-GPT-2.7B  &38.60  &20.60  &24.60    &27.93\\
\myrowcolour
\textbf{SmolLM-1.7B $+$ Ours} &\textbf{48.60}  &\textbf{31.60}  &\textbf{33.00}    & \textbf{37.73}\\
\bottomrule
\end{tabular}%
}
\end{table}

Table \ref{tab:baselines} shows that our method consistently delivers the best accuracy–latency trade-off across all SmolLM scales, achieving near-FP16 accuracy while significantly reducing decoding cost relative to static and heuristic KV quantization baselines. Table \ref{tab:models} further confirms the effectiveness of the proposed framework, as SmolLM + Ours attains the strongest overall performance across all datasets.

\section{Conclusion}

This paper presented \textit{\textbf{Don’t Waste Bits!}}, an adaptive KV-cache quantization framework for efficient LLM inference. By combining lightweight token-level saliency features with a compact controller, the proposed method dynamically assigns KV precision during decoding, enabling heterogeneous bit-width allocation across tokens. This design improves the trade-off among predictive accuracy and decoding latency across multiple benchmarks and model scales. The framework is also practically deployable, since its features are inexpensive to compute, the controller adds minimal overhead, and precision decisions are produced online without costly search, iterative optimization, or modifications to the transformer. As a result, the method integrates naturally into standard inference pipelines and supports memory-efficient, latency-aware LLM deployment.


{
    \small
    \bibliographystyle{ieeenat_fullname}
    \bibliography{main}
}


\end{document}